\documentclass[runningheads]{llncs}
\usepackage[T1]{fontenc}
\usepackage{graphicx}
\usepackage{booktabs}
\usepackage[misc]{ifsym}
\newcommand{\corr}{(\Letter)}
\usepackage{graphicx,appendix,float}
\usepackage{url,amsmath,amssymb,fancybox,listings,pdfpages,caption,multicol,datetime,rotating,booktabs}
\usepackage{algorithm,algpseudocode}
\usepackage{multirow}
\usepackage{subcaption}
\usepackage{pifont}
\usepackage{nicematrix}
\usepackage{tablefootnote}
\usepackage{threeparttable}
\usepackage{tabularx}
\usepackage{ragged2e}
\newcolumntype{L}{>{\RaggedRight\arraybackslash}X}
\newcolumntype{S}{>{\RaggedRight\arraybackslash\hsize=.6\hsize}X} 

\begin{document}

\title{Relative Parameter Importance in Task-Agnostic Replay-Free Continual Learning}

\titlerunning{Relative Parameter Importance in Continual Learning}

\author{Malavika Suresh\inst{1} \corr \and
Ikechukwu Nkisi-Orji\inst{1} \and
Nirmalie Wiratunga\inst{1}}

\authorrunning{M. Suresh et al.}

\institute{Robert Gordon University \email{m.suresh@rgu.ac.uk}
}

\maketitle              

\begin{abstract}
Achieving continual learning (CL) with deep neural networks requires balancing stability and plasticity while enabling knowledge transfer. In this work, we focus on offline learning algorithms under the constraints: (I) no access to training data from prior tasks (II) no access to task-id at inference time. We introduce a novel measure, the \textit{relative parameter-importance}, which measures the relative importance of each parameter with respect to both the current and past tasks. Parameters with high relative importance are interpreted as more important for maintaining past-task stability and thus heavily regularised, whereas parameters with low relative-importance are allowed to be more freely updated. Unlike existing methods, our approach allows the update of parameters with high past-task importance when they have low relative-importance, thus enabling backward knowledge transfer in addition to tackling the stability-plasticity trade-off. We demonstrate improvements against state-of-the-art CL methods on both class-incremental and domain-incremental learning text classification problems\footnote{Code available at: \url{https://github.com/itsmemala/LACL}} and provide insights for extending our method to text generation problems.

\end{abstract}                

\section{Introduction}
Avoiding the Catastrophic Forgetting (CF) of learnt abilities when training on new data (i.e. the stability-plasticity trade-off) and enabling Knowledge Transfer (KT) between old and new data are crucial elements to bridging the gap between Continual Learning (CL) and Multi-Task Learning (MTL) in deep neural networks.
\textit{Experience replay} \cite{dautumn2019mbpa,buzzega2020der++} and \textit{parameter masking} \cite{serra2018hat,ke2021ctr} are two common CL paradigms
but they pose practical challenges. Replay requires storing data from prior tasks which incurs memory costs and violates data privacy, while parameter masking requires knowing which task (i.e. a task-id) the inference sample originates from, which is often practically infeasible.
In this work, we focus on the challenging setting of CL with (I) no access to past-task data (replay-free) and (II) no access to task-id (task-agnostic) at inference time.

Traditional \textit{gradient projection} \cite{zeng2019owm,saha2021gpm} and \textit{regularisation} methods \cite{kirkpatrick2017ewc,Li2016LWF} 
tackle CL under constraints I and II but focus mainly on preventing CF. 
Gradient projection methods restrict gradient updates for new tasks so they do not overlap with directions important to past tasks (using orthogonality constraints),
while regularisation methods restrict changes to model parameters or layer representations deemed important for past tasks (using regularisation objectives). 
These constraints restrict model plasticity and KT, particularly when new tasks are correlated with past tasks.
To address this, some approaches relax the constraint based on task correlation \cite{lin2022trgp,cheng2025adabop}, or perform a joint regularisation of the main network towards both the frozen past-task model and an auxiliary model trained solely on the new task \cite{kim2023ancl}. 
The success of these methods, however, depends on selecting optimal orthogonality/regularisation hyper-parameters.
In the replay-free setting, it is difficult to estimate the impact of such hyper-parameters on past-task performance (i.e. knowledge transfer versus forgetting), which risks sub-optimal performance in practice.
To address this concern, we propose a granular parameter-level solution that identifies and enables the update of parameters that facilitate backward knowledge transfer while ensuring that parameters more likely to cause forgetting are regularised.

Our proposed method consists of two training phases for each new task introduced to the continual learner - (i) Look-Ahead (LA) Phase: A temporary auxiliary model is initialised with the frozen past-task model and trained solely on the new task data, after which importance estimates for each model parameter are derived using a novel notion of \textbf{relative-importance}. (ii) Main Continual Learning (MCL) Phase: The main continual learner is trained with a relative-importance based regularisation objective. 
Unlike the traditional parameter regularisation methods \cite{kirkpatrick2017ewc,aljundi2018mas,chaudhry2018CLintransigence}, that only estimate the importance with respect to past tasks, here we define the \textbf{relative-importance} of a parameter as the ratio of the past-task importance to the sum of the past-task importance and current-task importance. 
Parameters with high relative-importance can thus be interpreted as more relevant for maintaining past-task stability and therefore updates to these parameters can to be regularised to prevent CF. Parameters with low relative-importance, on the other hand, can be interpreted as contributing to the learning of new task as well as shared knowledge and can be less regularised for allowing plasticity and backward knowledge transfer. 
Our goal is to perform an \textit{informed} regularisation of parameters in order to improve performance.

Lastly, the rise in the use of pre-trained generative models has led to the emergence of new CL areas such as \textit{continual pre-training} \cite{abbes2025continualpretraining} and \textit{continual fine-tuning} to add new abilities \cite{sanyal2025upweighting} and ensure alignment \cite{lu2024onlinemergingoptimizersboosting}. Research towards employing traditional CL techniques to generative models is still naive. Moreover, lack of access to pre-training data and the large overlap in core language/vision abilities across different tasks make replay, parameter-masking and gradient projection difficult to implement. In this work, we analyse how parameter importance based regularisation can be extended to generative language models and provide insights to help drive research in this direction. 

In summary, we make the following contributions:
\begin{enumerate}
    \item We introduce a novel relative parameter-importance measure for replay-free, task-id-free continual learning, estimating each parameter's significance for both current and past tasks to enable informed regularisation.
    \item We evaluate our proposed approach against state-of-the-art CL methods and show that our method leads to improved performance on both domain-incremental and class-incremental learning text classification scenarios.
    \item We discuss insights for extending our method to text generation scenarios.
\end{enumerate}


\section{Background and Related Work}
\label{section:rw}
\textbf{Replay} methods store some data from past tasks for re-use to mitigate CF \cite{dautumn2019mbpa,lin2024DGC}. This incurs high memory costs over several tasks
and also violates data privacy. 
While some propose using generative models for synthetic replay \cite{sun2020lamol}, or using activations for replay \cite{sarfraz2025sarl}, it is hard to ensure
generation accuracy
and to apply similarity constraints on activations when old and new task domains contain same classes.
\textbf{Parameter masking} approaches \cite{serra2018hat,kang2022wsn,hu2024osn} 
use masks during forward/backward propagation to prevent re-use of parameters important to past tasks,
resulting in task-specific sub-networks.
Some methods \cite{ke2021kan,ke2021ctr}
also allow parameter sharing between tasks.
In all these methods, to identify the right sub-network at test time, a task-id must be selected. This trivializes the problem by identifying a subset of classes rather than classifying among all seen classes \cite{chaudhry2018CLintransigence}. 
While some methods \cite{wortsman2020supsup,rajasegaran2020itaml,Dekhovich2023cps} predict a task-id, they require a batch making them impractical when only a single test sample is presented.
\textbf{Gradient projection} methods \cite{lopez2017gem,zeng2019owm,saha2021gpm} ensure that updates from a new task are orthogonal to the subspace spanned by previous task inputs to minimise CF.
However, this affects KT and plasticity as parameter updates are increasingly restricted to directions orthogonal to all previous tasks. Though several methods \cite{lin2022trgp,yang2025flatnessawareOGP,cheng2025adabop}
propose to tackle this by relaxing the orthogonality constraint, 
they can still result in sub-optimal learning when there is high overlap between tasks, as shown in our experiments.
\textbf{Representation regularisation} methods \cite{Li2016LWF,Szatkowski2024TAKD} employ regularisation objectives to tackle CF by preventing change in the model representations corresponding to old tasks when learning the new task. 
To enable KT, some works \cite{ebrahimi2020ACL,luo2024accl} decompose the representation space into shared and task-specific features and use additional objectives to encourage similarity between shared features.  Extracting task-specific features at test-time, however, requires a task-id.

\textbf{Parameter importance based regularisation} methods estimate the importance of each parameter \cite{kirkpatrick2017ewc,aljundi2018mas,chaudhry2018CLintransigence} or network node \cite{ahn2019ucl,jung2020agscl} 
and regularise the learning rate for parameters important to past tasks 
to minimise CF. 
Some methods 
directly fuse past-task and current-task parameters with importance-based weighting \cite{sun2024RP2F}, or perform importance-based gradient updates
\cite{elsayed2024UPGD}. In all these methods, parameters with high past-task importance are \textit{always} prevented from being updated, which over-constrains the network and affects plasticity and KT. Our proposed method tackles this issue by introducing the notion of \textit{relative-importance}.
Unlike other auxiliary model based methods such as \cite{kim2023ancl}, that employ two regularisation terms, we derive a single importance estimate for each parameter and use only \textit{one} regularisation term. This enables a more \textit{informed} regularisation and avoids the complexity of balancing multiple objectives.

\section{Methodology}
\label{section:method}

\paragraph{Notations} Consider N sequential tasks $\{ T_{1}, T_{2}, ..., T_{N} \}$. Each task $T_k$ consists of a train $tr_k$, a validation $v_k$ and a test $t_k$ split. We denote the model trained on $T_k$ as $M_k$ and the model parameters as $\theta_k$. The goal is to train $M_k$ while retaining its performance on tasks $\{T_{1},...,T_{k-1}\}$ without access to their data. 
$M_k^{la}$ denotes the model at the end of the look-ahead phase for the $k^{th}$ task.

\subsection{Overview: Continual Learning with Look-Ahead (LA)}\label{section:method_overview}

\begin{figure*}[t]
\centering
\includegraphics[width=1.0\textwidth]{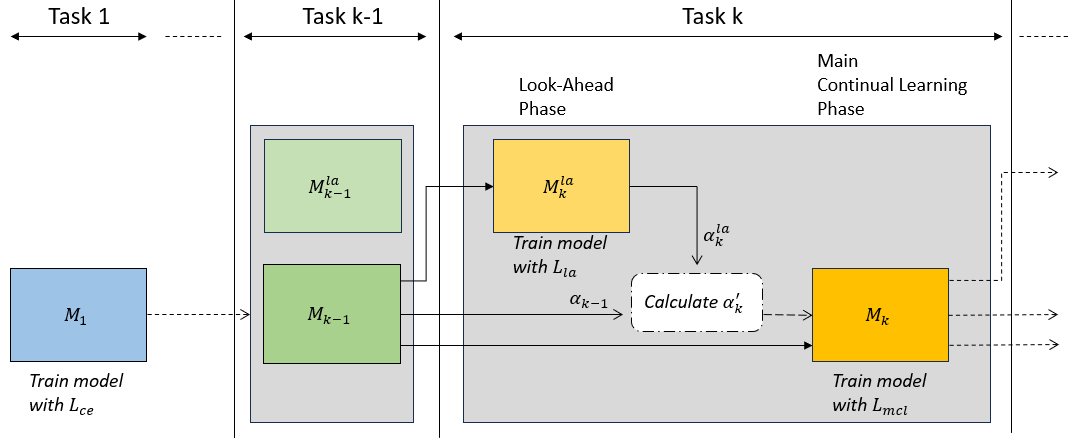}
\caption{Look-Ahead (LA) continual learning. The model $M_1$ is trained on the first task using a cross-entropy loss. For each subsequent task $k$, model training is done in two phases - (i) Look-Ahead Phase: A temporary model $M_k^{la}$ is initialised using $M_{k-1}$ and trained solely on the new task with no regularisation.
At the end of this phase, the relative importance based regularisation weights ($\alpha_k'$) are estimated. (ii) Main Continual Learning Phase: The main model $M_k$ is initialised using $M_{k-1}$ and trained with relative importance based regularisation.}
\label{fig:method_overview}
\end{figure*}

Figure \ref{fig:method_overview} depicts the training process. We assume a fixed capacity model. 
Each task is trained using a task-specific cross-entropy loss ($L_{ce}$). 
At the end of the training for each task $k$, the importance of each parameter $j$ is estimated with respect to that task, denoted as $\alpha_{k,j}$. Here, we use the method of Memory Aware Synapses (MAS) \cite{aljundi2018mas}, which computes the importance as the sensitivity of the learned output function to the parameter, as shown in Equation \ref{eqn:mas}, i.e. the gradient when back-propagating the $L_2$-norm of the output logits.

\begin{equation}
\label{eqn:mas}
    \alpha_{k,j} = \frac{\partial M_k(tr_k)}{\partial \theta_{k,j}}
\end{equation}

For simplicity, $\alpha_k$ is used to denote the set of all parameter importance values $\{\alpha_{k,j}\}$ for the $k^{th}$ task. The importance values are aggregated across tasks (to avoid memory cost of storing values for each task) by mean-pooling, such that $\alpha_k = mean\{\alpha_0,\alpha_1,...,\alpha_k\}$. This ensures that parameters that continue to be important for multiple tasks retain higher importance values and are better protected through importance-based regularisation.\footnote{It is noted that mean-pooling leads to loss of some granularity. While max-pooling is an alternative, we believe it could cause early capacity saturation through high importance for most parameters  - we leave further exploration to future work.}



For the second task onwards, training is conducted in two phases - the look-ahead training ($la$) phase and the main continual learning ($mcl$) phase. During the look-ahead phase, the model is optimised solely on the new task without any regularisation using Equation \ref{eqn:la_loss}.

\begin{equation}
    \label{eqn:la_loss}
    L_{la} = L_{task} = L_{ce}
\end{equation}

The look-ahead training phase acts as a performance baseline and also identifies the gradient update path taken by the new task. At the end of this phase, the importance of parameters with respect to the new task are estimated and used to obtain the relative importance of each parameter with respect to past and current tasks. This helps inform which parameters are likely to cause CF and which may help with KT. Based on this, a modified weight $\alpha'$ is calculated, which allows parameters likely to help with KT to be easily updated, even if they were of high importance to past tasks. During the main continual learning phase, this modified weight is used in the regularisation, as shown in equation \ref{eqn:mcl_loss}, to selectively control the parameter plasticity. The regularisation in this phase thus focuses on balancing CF and KT, as opposed to merely avoiding CF. The strength of the regularisation is controlled using the hyper-parameter $\lambda$. 

\begin{equation}
\label{eqn:mcl_loss}
    L_{mcl} = L_{task} + \dfrac{\lambda}{2}\sum_j\alpha'_{k,j}(\theta_{k-1,j}-\theta_{k,j})^2
\end{equation}

At the end of the main continual learning phase, the latest parameter importance with respect to the new task $\alpha_k$ is calculated and mean-pooled for use in the next task regularisation. Next, we discuss how the parameter importance values for the main continual learning phase $\alpha_k'$ are estimated.

\subsection{Controlling Parameter Plasticity using Relative-Importance}
Let $\alpha_k^{la}$ denote the importance of parameters with respect to the new task $k$ at the end of the look-ahead training phase. The relative importance of parameters $\alpha_k^{rel}$ is computed as shown in equation \ref{eqn:alpha_rel} and is in the range $[0,1]$. $\epsilon$ is a small constant to ensure numerical stability, i.e. to ensure $\alpha_k^{rel}=0$ when both $\alpha_{k-1}=0$ and $\alpha_k^{la}=0$.\footnote{Since $\alpha$ is computed as the gradient with respect to the parameter, based on the floating point precision used in practice, it can sometimes be zero. We only observe this for a negligible number of parameters in our experiments and set $\epsilon=1e^{-10}$ to ensure  its effect on the denominator is otherwise negligible.}

\begin{equation}
\label{eqn:alpha_rel}
    \alpha_k^{rel}=\dfrac{\alpha_{k-1}}{\alpha_{k-1}+\alpha_k^{la}+\epsilon}
\end{equation}

Here, we interpret parameters with a high $\alpha_k^{rel}$ value as more important to preserving past task knowledge than for learning new task knowledge. Therefore, changes to these parameters must be minimal to help prevent CF. On the other hand, parameters with a low $\alpha_k^{rel}$ value are interpreted as more important to new task knowledge accumulation. Therefore, by allowing these parameters to be updated, we can maintain model plasticity, while potentially also allowing backward KT. 

The new constraint $\alpha_k'$ for each parameter is determined by weighting the original $\alpha_{k-1}$ value using $\alpha_k^{rel}$, as shown in Equation \ref{eqn:alpha_kdash}, where $\tau_{\alpha^{rel}}$ is the cut-off for differentiating high and low relative-importance and  $\lambda_{up}$ and $\lambda_{down}$ are scaling hyper-parameters.
The first term focusses on CF avoidance, where the constraint on parameters with high relative-importance is increased through the weight $\lambda_{up}.\alpha_k^{rel}$. $\lambda_{up}$ is lower bounded by $1/\tau_{\alpha^{rel}}$ to ensure $\lambda_{up}.\alpha_k^{rel}>1$ (to mitigate CF). The second term focuses on plasticity and KT, where the constraint on parameters with low relative-importance is decreased through the weight $\lambda_{down}.\alpha_k^{rel}$. $\lambda_{down}$ is upper bounded by 1 to ensure $\lambda_{down}.\alpha_k^{rel}<1$ (to enable plasticity/KT). See Appendix \ref{section:analysis} for a detailed empirical analysis of alternative weighting strategies considered.

\begin{equation}
\label{eqn:alpha_kdash}
    \alpha_k' =
    \begin{cases}
        \lambda_{up}.\alpha_k^{rel}.\alpha_{k-1} & \text{if $\alpha_k^{rel}>\tau_{\alpha^{rel}}$} \\
        \lambda_{down}.\alpha_k^{rel}.\alpha_{k-1} & \text{otherwise}
    \end{cases}
\end{equation}

The importance weight in quadratic regularisers can be seen as regulating the learning rate for each parameter \cite{lubana2021ReguInstability,kim2023ancl}. Our approach adjusts this learning rate by taking into account the relative importance of a parameter for current vs past tasks (through $\alpha_k^{rel}$).
Algorithm \ref{alg:la-with-chsf} (Appendix \ref{ap:alg}) summarises the look-ahead method, including a replay-free hyper-parameter search procedure.

\section{Text Classification Experiments}
This section evaluates the proposed Look-Ahead method on text classification.
\subsection{Experiment Setup}\label{section:setup}
\subsubsection{Datasets}

\paragraph{1. Intent classification (CIL)} This involves classifying home assistant queries \cite{liu2021hwu64} by their intent type (e.g. `set alarm', `play music'). The most frequent intents are split to form five tasks of five intents each. This dataset models the CIL setting, which is more challenging when not using a task-id based method \cite{chaudhry2018CLintransigence}. 

\paragraph{2. Behaviour classification (DIL)} This involves classifying client behaviour (as change/neutral/sustain) in motivational interviewing (MI) conversations between a client and a therapist \cite{wu2022annomi} in different domains. This task is important for automating MI conversations and evaluating MI effectiveness \cite{tavabi2021analysisMIcodes}. Data collection often involves focus-groups of one domain (e.g. alcohol/anxiety) 
with high label annotation cost and privacy concerns preventing data sharing. This motivates a CL approach. We use the data from \cite{wu2022annomi} to form six domains based on the conversation topic (see Appendix \ref{ap:behav-classification}). The dataset forms an interesting challenge due to in-domain class imbalance and cross-domain sample imbalance. 

Class-wise examples of each dataset are provided in Table \ref{tab:cl-data-examples} in Appendix \ref{ap:dataset-examples}.

\subsubsection{Performance Metrics}
Macro-averaged F1 scores on the test sets are used to measure performance on individual tasks. Let $F_{k,i}$ denote the F1 score of the model $M_k$ on task i. The following metrics are used to report CL performance:

\begin{enumerate}
    \item Overall ($Ov$): Average performance across all tasks at the end of sequence.
    \begin{equation}
        \dfrac{1}{N} \sum_{i=1}^{N} F_{N,i}
    \end{equation}
    \item Catastrophic Forgetting ($CF$): Average performance loss 
    on previous tasks after sequentially learning all tasks \cite{huang2021idbr}, i.e. for each task $i$, the performance difference between the final model $N$ and the previous best model $k$. 
    \begin{equation}
        \dfrac{1}{N-1} \sum_{i=1}^{N-1}[
        {\max_{i \leq k \leq N-1}(F_{k,i})-
        F_{N,i}}
        ]
    \end{equation}
    \item Positive backward transfer ($BWT^+$): Average gain in performance on seen tasks after learning each new task \cite{diaz2018CLeval}, i.e. for each task $i$, the gain in performance when going from the previous ($k-1$) to the current model ($k$).
    \begin{equation}
        \dfrac{1}{(\dfrac{N(N-1)}{2})} \sum_{k=2}^{N}
        \sum_{i=1}^{k-1}
        {\max(0,F_{k,i}-
        F_{k-1,i}})
    \end{equation}
    \item Forward transfer ($FWT$): Average gain in performance on each task when it is first introduced to the CL model, relative to naive sequential learning ($seq$) \cite{ke2021kan}.
    Positive values imply good model plasticity and KT from prior tasks to current task, while negative values typically imply loss of plasticity.
    \begin{equation}
        \dfrac{1}{N-1} \sum_{i=2}^{N} [F_{i,i}-F_{i,i}^{seq}]
    \end{equation}
\end{enumerate}

\subsubsection{Compared Methods}
Our main baselines are \textbf{AdaBOP} \cite{cheng2025adabop} (task-correlation based gradient projection), \textbf{UPGD} \cite{elsayed2024UPGD} (parameter regularisation), and \textbf{RP2F} \cite{sun2024RP2F} and \textbf{ANCL} \cite{kim2023ancl} (auxiliary network based regularisation). We apply ANCL to parameter regularisation \textbf{MAS} \cite{aljundi2018mas} and representation regularisation \textbf{LWF} \cite{Li2016LWF}. 
We report multi-tasked \textbf{MTL} and naive sequential learning \textbf{SEQ} baselines. 

\subsubsection{Implementation Details}
The network backbone is Adapter-BERT \cite{houlsby2019bertadapter}, where only the adapters, layer-norm and classifier head parameters are trainable during CL.
Hyper-parameter search is implemented with $drop=10$, $thr=90$ for CIL and $thr=95$ for DIL (per algorithm \ref{alg:la-with-chsf}). $\tau_{\alpha^{rel}}$ is dynamically set to 80\% of the $\alpha^{rel}$ distribution mean at each model component and layer (per analysis in Appendix \ref{section:analysis} and Appendix \ref{ap:ablation-tau-alpha-rel}). Further details (task splits, etc.) are in Appendix \ref{ap:imp-details}.

\subsection{Results and Discussion}

\begin{table*}[h]
\caption{Mean (std) performance at the end of all tasks, measured in F1 scores and averaged across three random task orderings. $\uparrow$ denotes higher is better, $\downarrow$ denotes lower is better. Best is indicated in bold, second best is underlined.}
\label{tab:CIL-main}
\begin{center}
\begin{tabular}{clllll}
\toprule
{Data} &{Method} &{ Ov$\uparrow$} &{ CF$\downarrow$} &{ BWT+$\uparrow$} &{ FWT$\uparrow$}
\\
\midrule
\multirow{8}{*}{CIL} &MTL &93.94 (2.4) &- &- &- \\
 &SEQ &15.22 (1.2) &79.20 (9.4) &0.00 (0.0) &- \\
 &AdaBOP &12.96 (6.2) &\textbf{62.30 (15.3)} &0.00 (0.0) &-20.01 (23.6) \\
 &RP2F &14.01 (9.9) &83.34 (3.7) &0.00 (0.0) &2.35 (4.7) \\
 &UPGD &29.58 (4.7) &74.73 (1.9) &0.56 (0.8) &\underline{13.13 (8.2)} \\
 &ANCL-MAS &\underline{31.19 (6.4)} &71.86 (4.6) &1.29 (0.7) &12.05 (8.6) \\
 &ANCL-LWF &21.64 (3.6) &85.62 (6.2) &\underline{1.72 (2.4)} &\textbf{13.88 (7.7)} \\
 &LA-MAS &\textbf{34.20 (7.4)} &\underline{67.44 (8.8)} &\textbf{7.88 (11.1)} &11.37 (9.2) \\
\midrule
\multirow{8}{*}{DIL} &MTL &54.72 (2.7) &- &- &- \\
 &SEQ &45.26 (3.5) &11.86 (5.8) &7.01 (1.7) &- \\
 &AdaBOP &44.1 (8.7) &7.12 (6.9) &3.68 (1.5) &-7.18 (6.1) \\
 &RP2F &46.83 (5.8) &7.99 (6.5) &4.83 (1.1) &\textbf{-2.52 (4.9)} \\
 &UPGD &42.57 (0.9) &\underline{3.86 (5.0)} &\underline{5.98 (2.9)} &-8.85 (1.7) \\
 &ANCL-MAS  &43.90 (1.9) &7.33 (3.4) &4.36 (0.5) &-7.77 (2.3)  \\ 
 &ANCL-LWF &\underline{50.02 (4.3)} &4.08 (3.4) &\textbf{7.61 (2.2)} &\underline{-2.84 (1.7)}  \\
 &LA-MAS &\textbf{50.48 (4.8)} &\textbf{2.64 (1.5)} &5.64 (1.0) &-3.98 (1.5) \\
 \bottomrule
\end{tabular}
\end{center}
\end{table*}


Table \ref{tab:CIL-main} compares our Look-Ahead (LA) approach to various CL baselines.
On the CIL problem, while some CL methods provide gains over the SEQ baseline, they all fall significantly short of the MTL upper bound, highlighting the difficulty of 
replay-free and task-agnostic
CIL. With the gradient projection based method AdaBOP, model plasticity (FWT) is significantly affected by the orthogonality constraint resulting in poor overall performance (Ov) even though forgetting (CF) is reduced on average compared to SEQ. RP2F (auxilliary network based method) improves plasticity slightly on average compared to SEQ but incurs significant CF resulting in poor overall performance again. UPGD (parameter regularisation method) improves plasticity significantly, resulting in an improved overall performance, though CF still remains high. For ANCL-MAS and ANCL-LWF (auxilliary network based methods), since the regularisation hyper-parameters control the stability-plasticity trade-off, a desired level of plasticity can be specified and hyper-parameters selected accordingly to maximise stability. The desired plasticity is specified as a percentage threshold of the best achievable performance for the current task (i.e. naive fine-tuning of the main network on current task). At a high threshold of 90\%, both ANCL baselines achieve high plasticity comparable to UPGD, but ANCL-MAS achieves better CF avoidance leading to better overall performance.
Using the same threshold for hyper-parameter selection, the LA method reduces CF while also improving backward KT at a similar plasticity level, resulting in the best overall performance. 
Appendix \ref{ap:thresholds-analysis} compares the performance at different thresholds.

On the DIL problem, with significant shared knowledge across tasks, the gap between MTL and SEQ is reduced. SEQ already achieves good model plasticity and backward KT, with the CL methods mainly improving performance by minimising CF. Several methods (AdaBOP, UPGD, ANCL-MAS) reduce CF at the cost of loss to plasticity, resulting in poor overall performance. RP2F reduces CF without significant impact to plasticity, resulting in a small improvement on average over SEQ. ANCL-LWF reduces CF significantly while also maintaining good plasticity and backward KT, resulting in a good overall performance. The LA method 
reduces CF further to achieve the best overall performance. 

Appendix \ref{ap:comp-cost} compares the computation cost of the various CL methods.

\section{Preliminary Analysis for Text Generation}\label{section:text_gen_expts}
This section performs preliminary analysis for extending our method to generative models. 
We first extend the parameter importance computation defined for classification outputs in Section \ref{section:method_overview} to sequence outputs, as follows: The importance is computed as the gradient when back-propagating the $L_2$-norm of the output token logits, \textit{averaged across all output tokens of the sequence}. Our experiments (1) support the proposed importance estimation method and, (2) reveal insights for next steps in this direction.
\subsection{Experiment Setup}
We use the pre-trained model checkpoint for Llama-3.2-1B. Our target task is mathematical reasoning - we fine-tune on MetaMathQA \cite{yu2024metamath} and evaluate the performance on GSM8K \cite{cobbe2021gsm8k}. We consider Python programming (MBPP \cite{jacob2021mbpp}) as a core pre-trained ability to retain after fine-tuning on target task. We use the evaluation setup of \cite{sanyal2025upweighting} for reporting performance on both pre-trained and target tasks.
We use epochs=2 and batch size=4 for fine-tuning.

\subsection{Results and Discussion}
First, we perform naive fine-tuning on target task and measure the correlation between the \textit{change in parameter magnitude} and the \textit{parameter importance} with respect to the target task, across model layers. The high correlation values in Table \ref{tab:corr_mas_sft} indicate that our proposed importance estimation method works - parameters that have changed to adapt to the task are indeed assigned higher importance. 
We then compute parameter importance of the pre-trained model with respect to MBPP ($\alpha_j$) and use this for fine-tuning with regularisation to prevent CF. We set $\alpha^{'}_j=\alpha_j$ for preliminary analysis. 
Results in Table \ref{tab:llama_mas_results} indicate that 
the training is unstable
for a wide range of $\lambda$.
This suggests that to implement parameter regularisation in generative models, a crucial next step for research is to effectively balance the cross-entropy and regularisation objectives.

\begin{table*}[h]
\caption{Statistics of correlation ($\in[-1,1]$) between absolute change in parameter magnitude ($|\Delta\theta_j|$) and parameter importance values ($\alpha_j$), across model layers, after naive fine-tuning (FT) a pre-trained (PT) model for a target task.}
\label{tab:corr_mas_sft}
\begin{center}
\begin{tabular}{c|cc|cccc}
\toprule
 &\multicolumn{2}{c}{Target task accuracy} &\multicolumn{4}{c}{Correlation statistics ($|\Delta\theta_j|,\alpha_j$)} \\
 \midrule
Model &PT model &FT model &Min &Median &Mean &Max \\
\midrule
Llama-3.2-1B &6.07 &42.00 &0.40 &0.63 &0.65 &0.98 \\
\bottomrule
\end{tabular}
\end{center}
\end{table*} 

\begin{table*}[h]
\caption{Results of parameter-importance regularisation during fine-tuning.}
\label{tab:llama_mas_results}
\begin{center}
\begin{tabular}{c|c|c|c}
\toprule
Method &$\lambda$ &Pre-trained task accuracy &Target task accuracy \\
\midrule
Pre-trained &- &26.60 &6.07\\
Naive fine-tuning &- &18.80 &42.00\\
Regularisation &0.01 &0.00 &2.58\\
Regularisation &0.10 &0.00 &2.27\\
Regularisation &1.00 &0.00 &1.67\\
Regularisation &10.00 &0.00 &1.29\\
Regularisation &100.00 &0.00 &1.97\\
Regularisation &1000.00 &0.00 &1.67\\
\bottomrule
\end{tabular}
\end{center}
\end{table*}

\section{Conclusion}
\label{section:conclusion}
In this work, we proposed the Look-Ahead (LA) method, which implements a novel \textit{relative parameter-importance} measure for replay-free and task-agnostic CL. Our method estimates the importance of parameters with respect to both the current and past tasks. 
Unlike existing methods, we allow parameters with high past-task importance to be updated when they have low relative-importance. Experiments demonstrated improvements using the LA method compared to state-of-the-art baselines on both DIL and CIL text classification. 
Further analysis revealed insights for extending our method to text generation.
Finally, we note that our 
approach
can be used with any importance estimation method (for compute-accuracy trade-off), as well as with replay and task-id based methods.

\begin{credits}

\subsubsection{\discintname}
The authors have no competing interests to declare that are relevant to the content of this article. 
\end{credits}
%
%
%
\bibliographystyle{splncs04}
\bibliography{references}

@inproceedings{kim2023ANCL,
  title={Achieving a better stability-plasticity trade-off via auxiliary networks in continual learning},
  author={Kim, Sanghwan and Noci, Lorenzo and Orvieto, Antonio and Hofmann, Thomas},
  booktitle={Proceedings of the CVPR Conference},
  year={2023}
}

@InProceedings{kang2022wsn,
  title = 	 {Forget-free Continual Learning with Winning Subnetworks},
  author =       {Kang, Haeyong and Mina, Rusty John Lloyd and Madjid, Sultan Rizky Hikmawan and Yoon, Jaehong and Hasegawa-Johnson, Mark and Hwang, Sung Ju and Yoo, Chang D.},
  booktitle = 	 {ICML},
  year = 	 {2022},
}

@article{ke2021ctr,
  title={Achieving forgetting prevention and knowledge transfer in continual learning},
  author={Zixuan Ke and Bing Liu and Nianzu Ma and Hu X and Lei Shu},
  journal={Advances in Neural Information Processing Systems},
  volume={34},
  year={2021}
}

@inproceedings{Sun2020lamol,
  author    = {Fan{-}Keng Sun and
               Cheng{-}Hao Ho and
               Hung{-}Yi Lee},
  title     = {{LAMOL:} LAnguage MOdeling for Lifelong Language Learning},
  booktitle = {International Conference on Learning Representations},
  year      = {2020}
}

@inproceedings{huang2021idbr,
    title = "Continual Learning for Text Classification with Information Disentanglement Based Regularization",
    author = "Huang, Yufan  and
      Zhang, Yanzhe  and
      Chen, Jiaao  and
      Wang, Xuezhi  and
      Yang, Diyi",
    booktitle = "Proceedings of NAACL",
    year = "2021",
}

@inproceedings{ke2021kan,
  title={Continual learning with knowledge transfer for sentiment classification},
  author={Zixuan Ke and Bing Liu and Hao Wang and Lei Shu},
  booktitle={Proceedings of ECML-PKDD},
  year={2021},
}

@inproceedings{buzzega2020der++,
author = {Buzzega, Pietro and Boschini, Matteo and Porrello, Angelo and Abati, Davide and Calderara, Simone},
title = {Dark Experience for General Continual Learning: A Strong, Simple Baseline},
year = {2020},
booktitle = {Proceedings of the 34th NeurIPS Conference}
}

@inproceedings{houlsby2019bertadapter,
  title={Parameter-efficient transfer learning for NLP},
  author={Houlsby, Neil and Giurgiu, Andrei and Jastrzebski, Stanislaw and Morrone, Bruna and De Laroussilhe, Quentin and Gesmundo, Andrea and Attariyan, Mona and Gelly, Sylvain},
  booktitle={ICML},
  year={2019}
}

@article{kirkpatrick2017ewc,
  title={Overcoming catastrophic forgetting in neural networks},
  author={Kirkpatrick, James and Pascanu, Razvan and Rabinowitz, Neil and Veness, Joel and Desjardins, Guillaume and Rusu, Andrei A and Milan, Kieran and Quan, John and Ramalho, Tiago and Grabska-Barwinska, Agnieszka and others},
  journal={Proceedings of the national academy of sciences},
  volume={114},
  number={13},
  year={2017},
  publisher={National Acad Sciences}
}

@article{zeng2019owm,
  title={Continual learning of context-dependent processing in neural networks},
  author={Zeng, Guanxiong and Chen, Yang and Cui, Bo and Yu, Shan},
  journal={Nature Machine Intelligence},
  volume={1},
  number={8},
  year={2019},
  publisher={Nature Publishing Group UK London}
}

@inproceedings{wu2022annomi,
  title={Anno-mi: A dataset of expert-annotated counselling dialogues},
  author={Wu, Zixiu and Balloccu, Simone and Kumar, Vivek and Helaoui, Rim and Reiter, Ehud and Recupero, Diego Reforgiato and Riboni, Daniele},
  booktitle={IEEE International Conference on Acoustics, Speech and Signal Processing},
  year={2022}
}

@inproceedings{tavabi2021analysisMIcodes,
    title = "Analysis of Behavior Classification in Motivational Interviewing",
    author = "Tavabi, Leili  and
      Tran, Trang  and
      Stefanov, Kalin  and
      Borsari, Brian  and
      Woolley, Joshua  and
      Scherer, Stefan  and
      Soleymani, Mohammad",
    booktitle = "7th Workshop on Computational Linguistics and Clinical Psychology",
    year = "2021",
}

@inproceedings{dautumn2019mbpa,
 author = {de Masson d\textquotesingle Autume, Cyprien and Ruder, Sebastian and Kong, Lingpeng and Yogatama, Dani},
 booktitle = {Advances in Neural Information Processing Systems},
 title = {Episodic Memory in Lifelong Language Learning},
 year = {2019}
}

@inproceedings{diaz2018CLeval,
  title={Don't forget, there is more than forgetting: new metrics for Continual Learning},
  author={Diaz-Rodriguez, Natalia and Lomonaco, Vincenzo and Filliat, David and Maltoni, Davide},
  booktitle={Continual Learning Workshop at NeurIPS},
  year={2018}
}

@inproceedings{chaudhry2018CLintransigence,
  title={Riemannian walk for incremental learning: Understanding forgetting and intransigence},
  author={Chaudhry, Arslan and Dokania, Puneet K and Ajanthan, Thalaiyasingam and Torr, Philip HS},
  booktitle={Proceedings of ECCV},
  year={2018}
}

@inproceedings{aljundi2018mas,
  title={Memory aware synapses: Learning what (not) to forget},
  author={Aljundi, Rahaf and Babiloni, Francesca and Elhoseiny, Mohamed and Rohrbach, Marcus and Tuytelaars, Tinne},
  booktitle={Proceedings of the European conference on computer vision (ECCV)},
  year={2018}
}

@inproceedings{serra2018hat,
  title={Overcoming catastrophic forgetting with hard attention to the task},
  author={Serra, Joan and Suris, Didac and Miron, Marius and Karatzoglou, Alexandros},
  booktitle={ICML},
  year={2018}
}

@inproceedings{lubana2021ReguInstability,
  title={How do Quadratic Regularizers Prevent Catastrophic Forgetting: The Role of Interpolation},
  author={Ekdeep Singh Lubana and Puja Trivedi and Danai Koutra and Robert P. Dick},
  booktitle={CoLLAs},
  year={2021}
}

@article{JODELET2022RBS,
title = {Balanced softmax cross-entropy for incremental learning with and without memory},
journal = {Computer Vision and Image Understanding},
year = {2022},
author = {Quentin Jodelet and Xin Liu and Tsuyoshi Murata},
}

@inproceedings{wortsman2020supsup,
author = {Wortsman, Mitchell and Ramanujan, Vivek and Liu, Rosanne and Kembhavi, Aniruddha and Rastegari, Mohammad and Yosinski, Jason and Farhadi, Ali},
title = {Supermasks in Superposition},
year = {2020},
booktitle = {NeurIPS}
}

@inproceedings{rajasegaran2020itaml,
  title={Itaml: An incremental task-agnostic meta-learning approach},
  author={Rajasegaran, Jathushan and Khan, Salman and Hayat, Munawar and Khan, Fahad Shahbaz and Shah, Mubarak},
  booktitle={Proceedings of the CVPR Conference},
  year={2020}
}

@article{Dekhovich2023cps,
	year = 2023,
publisher = {Springer Science and Business Media {LLC}
},
author = {Aleksandr Dekhovich and David M.J. Tax and Marcel H.F Sluiter and Miguel A. Bessa},
title = {Continual prune-and-select: class-incremental learning with specialized subnetworks},
journal = {Applied Intelligence}
}

@article{lopez2017gem,
  title={Gradient episodic memory for continual learning},
  author={Lopez-Paz, David and Ranzato, Marc'Aurelio},
  journal={Advances in neural information processing systems},
  volume={30},
  year={2017}
}

@ARTICLE{luo2024accl,
  author={Luo, Yuxuan and Cong, Runmin and Liu, Xialei and Ip, Horace Ho Shing and Kwong, Sam},
  journal={IEEE Transactions on Multimedia}, 
  title={Modeling Inner- and Cross-Task Contrastive Relations for Continual Image Classification}, 
  year={2024}
  }

@InProceedings{Szatkowski2024TAKD,
    author    = {Szatkowski, Filip and Pyla, Mateusz and Przewi\k{e}\'zlikowski, Marcin and Cygert, Sebastian and Twardowski, Bart{\l}omiej and Trzci\'nski, Tomasz},
    title     = {Adapt Your Teacher: Improving Knowledge Distillation for Exemplar-Free Continual Learning},
    booktitle = {Proceedings of the WACV Conference},
    year      = {2024},
}

@inproceedings{elsayed2024UPGD,
  author={Mohamed Elsayed and A. Rupam Mahmood},
  title={Addressing Loss of Plasticity and Catastrophic Forgetting in Continual Learning},
  year={2024},
  booktitle={ICLR}
}

@inproceedings{sun2024RP2F,
author = {Sun, Wenju and Li, Qingyong and Zhang, Siyu and Wang, Wen and Geng, Yangliao},
title = {Incremental Learning via Robust Parameter Posterior Fusion},
year = {2024},
booktitle = {Proceedings of the ACM International Conference on Multimedia},
}

@inproceedings{lin2024DGC,
author = {Lin, Weichen and Chen, Jiaxiang and Huang, Ruomin and Ding, Hu},
title = {An effective dynamic gradient calibration method for continual learning},
year = {2024},
booktitle = {Proceedings of the 41st ICML},
}

@inproceedings{
saha2021gpm,
title={Gradient Projection Memory for Continual Learning},
author={Gobinda Saha and Isha Garg and Kaushik Roy},
booktitle={International Conference on Learning Representations},
year={2021}
}

@inproceedings{
lin2022trgp,
title={{TRGP}: Trust Region Gradient Projection for Continual Learning},
author={Sen Lin and Li Yang and Deliang Fan and Junshan Zhang},
booktitle={International Conference on Learning Representations},
year={2022}
}

@article{Li2016LWF,
  title={Learning without Forgetting},
  author={Zhizhong Li and Derek Hoiem},
  journal={IEEE TPAMI},
  year={2016}
}

@inproceedings{ebrahimi2020ACL,
  title={Adversarial continual learning},
  author={Ebrahimi, Sayna and Meier, Franziska and Calandra, Roberto and Darrell, Trevor and Rohrbach, Marcus},
  booktitle={Proceedings of ECCV},
  year={2020}
}

@inproceedings{liu2021hwu64,
  title={Benchmarking natural language understanding services for building conversational agents},
  author={Liu, Xingkun and Eshghi, Arash and Swietojanski, Pawel and Rieser, Verena},
  booktitle={Int. Workshop on Spoken Dialogue Systems},
  year={2021}
}

@article{lange2019chsf,
  author       = {Matthias De Lange and
                  Rahaf Aljundi and
                  Marc Masana and
                  Sarah Parisot and
                  Xu Jia and
                  Ales Leonardis and
                  Gregory G. Slabaugh and
                  Tinne Tuytelaars},
  title        = {Continual learning: A comparative study on how to defy forgetting in classification tasks},
  journal      = {CoRR},
  year         = {2019},
}

@inproceedings{
sarfraz2025sarl,
title={Semantic Aware Representation Learning for Lifelong Learning},
author={Fahad Sarfraz and Elahe Arani and Bahram Zonooz},
booktitle={International Conference on Learning Representations},
year={2025},
}

@inproceedings{hu2024osn,
author = {Hu, Yusong and Cheng, De and Zhang, Dingwen and Wang, Nannan and Liu, Tongliang and Gao, Xinbo},
title = {Task-aware orthogonal sparse network for exploring shared knowledge in continual learning},
year = {2024},
booktitle = {ICML}
}

@ARTICLE{cheng2025adabop,
  author={Cheng, De and Hu, Yusong and Wang, Nannan and Zhang, Dingwen and Gao, Xinbo},
  journal={IEEE Transactions on Circuits and Systems for Video Technology}, 
  title={Achieving Plasticity-Stability Trade-off in Continual Learning Through Adaptive Orthogonal Projection}, 
  year={2025}}

@ARTICLE{yang2025flatnessawareOGP,
  author={Yang, Enneng and Shen, Li and Wang, Zhenyi and Liu, Shiwei and Guo, Guibing and Wang, Xingwei and Tao, Dacheng},
  journal={IEEE TPAMI}, 
  title={Revisiting Flatness-Aware Optimization in Continual Learning With Orthogonal Gradient Projection}, 
  year={2025},
  }

@inproceedings{ahn2019ucl,
author = {Ahn, Hongjoon and Cha, Sungmin and Lee, Donggyu and Moon, Taesup},
title = {Uncertainty-based continual learning with adaptive regularization},
year = {2019},
booktitle = {Proceedings of the 33rd NeurIPS Conference}
}

@inproceedings{jung2020agscl,
 author = {Jung, Sangwon and Ahn, Hongjoon and Cha, Sungmin and Moon, Taesup},
 booktitle = {Proceedings of the NeurIPS Conference},
 title = {Continual Learning with Node-Importance based Adaptive Group Sparse Regularization},
 year = {2020}
}

@inproceedings{
yu2024metamath,
title={MetaMath: Bootstrap Your Own Mathematical Questions for Large Language Models},
author={Longhui Yu and Weisen Jiang and Han Shi and Jincheng YU and Zhengying Liu and Yu Zhang and James Kwok and Zhenguo Li and Adrian Weller and Weiyang Liu},
booktitle={International Conference on Learning Representations},
year={2024},
}

@article{cobbe2021gsm8k,
  title={Training Verifiers to Solve Math Word Problems},
  author={Karl Cobbe and Vineet Kosaraju and Mo Bavarian and Mark Chen and Heewoo Jun and Lukasz Kaiser and Matthias Plappert and Jerry Tworek and Jacob Hilton and Reiichiro Nakano and Christopher Hesse and John Schulman},
  journal={ArXiv},
  year={2021},
}

@article{jacob2021mbpp,
  author       = {Jacob Austin and
                  Augustus Odena and
                  Maxwell I. Nye and
                  Maarten Bosma and
                  Henryk Michalewski and
                  David Dohan and
                  Ellen Jiang and
                  Carrie J. Cai and
                  Michael Terry and
                  Quoc V. Le and
                  Charles Sutton},
  title        = {Program Synthesis with Large Language Models},
  journal      = {CoRR},
  year         = {2021}
}

@inproceedings{
abbes2025continualpretraining,
title={Revisiting Replay and Gradient Alignment For Continual Pretraining of Large Language Models},
author={Istabrak Abbes and Gopeshh Subbaraj and Matthew Riemer and Nizar Islah and Benjamin Th{\'e}rien and Tsuguchika Tabaru and Hiroaki Kingetsu and Sarath Chandar and Irina Rish},
booktitle={Women in Machine Learning Workshop at NeurIPS},
year={2026}
}

@inproceedings{
sanyal2025upweighting,
title={Upweighting Easy Samples in Fine-Tuning Mitigates Forgetting},
author={Sunny Sanyal and Hayden Prairie and Rudrajit Das and Ali Kavis and Sujay Sanghavi},
booktitle={International Conference on Machine Learning},
year={2025}
}

@misc{lu2024onlinemergingoptimizersboosting,
      title={Online Merging Optimizers for Boosting Rewards and Mitigating Tax in Alignment}, 
      author={Keming Lu and Bowen Yu and Fei Huang and Yang Fan and Runji Lin and Chang Zhou},
      year={2024},
      eprint={2405.17931},
      archivePrefix={arXiv},
      primaryClass={cs.CL},
      url={https://arxiv.org/abs/2405.17931}, 
}
%






\appendix
\section{Look Ahead Continual Learning with Continual Hyper-Parameter Search}
\label{ap:alg}
Selecting the optimal hyper-parameters without access to past-task data is challenging since the impact of the hyper-parameter choice on model stability cannot be directly observed. 
The LA approach minimises this challenge by identifying parameters with high relative importance as those that are most likely to impact model stability negatively. However, suitable hyper-parameter choices for $\lambda$, $\lambda_{up}$, $\lambda_{down}$ and $\tau_{\alpha^{rel}}$ still need to be made. In Appendix \ref{section:analysis}, a detailed analysis of hyper-parameter choices is conducted to understand the impact of different hyper-parameters on the continual learning performance. Here, an overall framework is described for selecting the hyper-parameters for each task in practice.

The proposed hyper-parameter selection approach is inspired by the Continual Hyper-parameter Search Framework (CHSF) \cite{lange2019chsf}. At the start of each task $k$, the best achievable performance for the task, $Acc_k$ (on the validation split $v_k$), is recorded using only cross-entropy and a grid search on the learning rate. Then, for $k>1$, hyper-parameter search for $\lambda$ is performed first by training with equation \ref{eqn:reg_loss}. 

\begin{equation}
\label{eqn:reg_loss}
    L_{reg} = L_{task} + \dfrac{\lambda}{2}\sum_j\alpha_{k-1,j}(\theta_{k-1,j}-\theta_{k,j})^2
\end{equation}

$\lambda$ is initialised to a high value such that current task performance=0\% and iteratively decreased by $drop\%$ until we reach at least threshold percentage ($thr\%$) of $Acc_k$ (on the validation split $v_k$). This ensures that the highest possible $\lambda$ is chosen, ensuring maximum stability at the given $thr\%$. Then, in order to select the LA specific hyper-parameters for $k>1$, the CHSF framework is extended as follows. To perform hyper-parameter search for $\lambda_{down}$ and $\lambda_{up}$ - we initialise first to their maximum values,
which ensures maximum stability, and decrease $\lambda_{down}$ first by $drop\%$ followed by $\lambda_{up}$ by $drop\%$, until we either improve the current task performance or reach a flat performance slope over $budget$ iterations. 
$\tau_{\alpha^{rel}}$ is set dynamically for each task based on the $\alpha^{rel}$ distribution (discussed in section \ref{section:analysis}). Following \cite{lange2019chsf}, $\lambda$ chosen from previous tasks are propagated to subsequent tasks. 

\paragraph{Bounding $\lambda_{up}$} Substituting $\alpha_k'$ from equations \ref{eqn:alpha_kdash} in equation \ref{eqn:mcl_loss}, we see that $\lambda_{up}$ and $\lambda_{down}$ act as scaling values for $\lambda$. While increasing $\lambda_{up}$ amounts to increasing the regularisation constraint on the corresponding parameters (i.e. those with high relative importance), in practice the values for $\lambda$ can vary widely based on the task, which can subsequently impact the choice of $\lambda_{up}$. Therefore, for simplicity, we derive an upper bound for $\lambda_{up}$ as $\lambda_{max}/\lambda$, where $\lambda_{max}$ is the largest value of $\lambda$ beyond which there is zero plasticity on the current task (using equation \ref{eqn:reg_loss}). The lower bound for $\lambda_{up}$ is given by $1/\tau_{\alpha^{rel}}$. Note that $\lambda_{max}$ relies only on access to current task data and can easily be determined using a coarse hyper-parameter search. 

The full Look-Ahead algorithm, including hyper-parameter search, is shown in algorithm \ref{alg:la-with-chsf}.

\begin{algorithm*}[h]
\caption{Continual Learning with Look-Ahead}
\label{alg:la-with-chsf}
\begin{algorithmic}[1]
\Require Model: $\theta_0$, Hyper-parameters: $thr, drop, \lambda_{init}, \tau_{\alpha^{rel}}$, $\{lr_1,lr_2,...\}$, $budget$, $\epsilon$.
\Require Accuracy estimation: $Acc()$, Parameter importance estimation: $Imp()$.
\For{task $k = 1, 2, \dots, N$}
    \For{$lr \in \{lr_1,lr_2,...\}$} \Comment{Grid search for learning rate}
        \State Initialise $\theta_k = \theta_{k-1}$.
        \State Train $\theta_k$ using $L_{task} = L_{ce}$ and $lr$. \Comment{No regularisation}
    \EndFor \Comment Record best accuracy $Acc_k$ and learning rate $lr_k$
    \If{$k = 1$}
        \State Initialise $\theta_k = \theta_{k-1}$.
        \State Train $\theta_k$ using $L_{task} = L_{ce}$ and $lr_k$. \Comment{No regularisation}
    \Else
        \State Initialise $\lambda = \lambda_{init}$.
        \While{$Acc(\theta_k,v_k) < thr.Acc_k$} \Comment{Search for $\lambda$}
            \State Initialise $\theta_k = \theta_{k-1}$.
            \State Train $\theta_k$ using eqn \ref{eqn:reg_loss} and $lr_k$.
            \State Update $\lambda=(1-drop).\lambda$.\Comment{Record $\lambda$ with Acc=0 as $\lambda_{max}$}
        \EndWhile \Comment Record accuracy with $\lambda$ as $Acc_\lambda$
        \State Initialise $\lambda_{init}=\lambda$, $\theta_k^{la} = \theta_{k-1}$.
        \State Train $\theta_k^{la}$ using $L_{task} = L_{ce}$ and $lr_k$.
        \Comment{LA Phase}
        \State Calculate $\alpha_k^{la}=Imp(\theta_k^{la})$ and $\alpha_k^{rel}$ using eqn \ref{eqn:alpha_rel}.
        \State Initialise $\lambda_{down}=1.0$, $\lambda_{up}=\lambda_{max}/\lambda$, $x=1$.
        \While{$Acc(\theta_k,v_k) < Acc_\lambda$ and $x<=budget$}  \Comment{Search for $\lambda_{down},\lambda_{up}$}
            \State Calculate $\alpha_k'$ using eqn \ref{eqn:alpha_kdash}.
            \State Train $\theta_k$ using eqn \ref{eqn:mcl_loss} and $lr_k$. \Comment{MCL Phase}
            \If{Slope($Acc(\theta_k,v_k)$) $<\epsilon$} \Comment{Performance plateaus}
                \State Update $\lambda_{up}=(1-drop).\lambda_{up}$.
            \Else
                \State Update $\lambda_{down}=(1-drop).\lambda_{down}$.
            \EndIf
            \State $x=x+1$
        \EndWhile
    \EndIf
    \State Calculate $\alpha_k=Imp(\theta_k)$ and $\alpha_k=meanpool(\alpha_k,\alpha_{k-1})$.
\EndFor
\end{algorithmic}
\end{algorithm*}

\section{Experiment Setup}
\subsection{Behaviour Classification: CL Domain Splits}\label{ap:behav-classification}
Table \ref{tab:domain-splits} shows the keywords used for segmenting the data into domains and the class-split in each domain.

\begin{table*}[t]
\caption{Domain Split}
\label{tab:domain-splits}
\begin{center}
\begin{tabular}{lcllll}
\toprule
\multicolumn{1}{c}{Domain}  &\multicolumn{1}{c}{Keywords} & &\multicolumn{3}{c}{Number of Samples}
\\ & & &Neutral &Change &Sustain
\\
\midrule
\multirow{2}{*}{Alcohol} &\multirow{2}{*}{`reducing alcohol'} &Train &457 &225 &77 \\
& &Test &115 &56 &19 \\
\hline
\multirow{2}{*}{Smoking} &\multirow{2}{*}{`smoking cessation'} &Train &232 &102 &82 \\
& &Test &58 &25 &21 \\
\hline
\multirow{2}{*}{Drug} &\multirow{2}{*}{`reducing drug use'} &Train &145 &61 &25 \\
& &Test &36 &16 &6 \\
\hline
\multirow{2}{*}{Exercise} &\multirow{2}{*}{`exercise',`weight loss'} &Train &185 &131 &22 \\
& &Test &46 &33 &6 \\
\hline
\multirow{2}{*}{Medicine} &\multirow{2}{*}{`medicine',`diabetes',`asthma'} &Train &345 &148 &93 \\
& &Test &87 &37 &23 \\
\hline
\multirow{2}{*}{Anxiety} &`anxiety management',`being assertive' &Train &80 &26 &4 \\
&`increasing self-confidence',`managing life' &Test &20 &7 &1
\\ \hline \\
\end{tabular}
\end{center}
\end{table*}


\subsection{Dataset Examples}\label{ap:dataset-examples}
Table \ref{tab:cl-data-examples} shows some examples from the two datasets.

\begin{table}[h]
\caption{Dataset Examples}
\label{tab:cl-data-examples}
\centering
\tiny
\begin{tabularx}{\linewidth}{@{} lll L LL @{}}
\toprule
Task & Domain & Intent/Class & Example & All Seen Domains & All Seen Intents/Classes \\ \midrule
\multicolumn{6}{c}{Intent Classification} \\ \midrule
\multirow{5}{*}{T1} & - & Music & play new gaga playlist. & \multirow{5}{*}{-} & \multirow{5}{*}{\{Music, Factoid, ...\}} \\
  & - & Music &Repeat the next song.\\
   & - & Factoid & Was Einstein married? & & \\ 
      & - & Factoid & what movies has neil walker done? & & \\ \cmidrule{2-4}   
\multirow{4}{*}{T2} & - & Remove & Erase all calendar events. & \multirow{4}{*}{-} &\{Music, Factoid, Remove, \\ 
      & - & Negate & please change that command. & &Negate, ...\} \\ 
    & - & Negate & that is not right answer. & & \\ 
\midrule
\multicolumn{6}{c}{Behaviour Classification} \\ \midrule
\multirow{5}{*}{T1} & Exercise & Change & I'm willing to start again, so to speak. & \multirow{5}{*}{\{Exercise\}} & \multirow{13}{*}{\{Change, Sustain, Neutral\}} \\
& Exercise & Sustain & Well, I don't see how I could do it. & & \\
& Exercise & Neutral & Yeah. & & \\ \cmidrule{2-4}
\multirow{8}{*}{T2} & Alcohol & Change & So, I could just tell them, `Hey, I don't feel like having a drink today.' & \multirow{8}{*}{\{Exercise, Alcohol\}} \\ 
& Alcohol & Sustain & And so, if I give up drinking, like, what am I going to
have to enjoy myself, you know. & & \\
& Alcohol & Neutral & Sounds good. & & \\
\bottomrule
\end{tabularx}
\end{table}

\subsection{Implementation Details}\label{ap:imp-details}
\paragraph{Compared methods implementation} We implement RP2F with fisher parameter importance, which is more computationally efficient and only marginally worse than using the parameter-perturbation based importance.\footnote{Based on results provided by authors.} For UPGD, we use the second-order utility function as proposed by authors.

\paragraph{Code}Code for UPGD, RP2F, AdaBOP and ANCL are adapted for text classification with the Adapter-BERT architecture using the description in the paper and code provided by the authors. 

\paragraph{Common training hyper-parameters} Across all methods, for each task we conduct a grid search on the learning rate$\in$\{0.00003, 0.0003, 0.003, 0.03\}. Models are trained for a maximum of 50 epochs with early stopping based on decrease in validation data loss. 


\paragraph{Hyper-parameter search for ANCL } ANCL requires setting two hyper-parameters - $\lambda$, which controls the strength of the regularisation towards the frozen past task model (i.e. controls stability), and $\lambda_a$, which controls the strength of the regularisation towards the auxiliary network trained on the new task (i.e. controls plasticity). Similar to the LA method, the CHSF framework is used where $\lambda$ is selected first using $drop=10$, followed by $\lambda_a$, which is initialised to $\lambda_a=0.01$ and increased by $inc=10$ until current task performance is improved. For comparison to the LA method, results are reported for the same thresholds.

\paragraph{Hyper-parameter search for RP2F} For selecting $\lambda$, which controls the strength of the parameter robustness term in the training objective function, a grid search $\in \{1e-5,1e-6,1e-7\}$ is performed for each task, as suggested by the authors, using the current task validation set performance for selection. Note that unlike the LA and ANCL methods, this hyper-parameter does not correlate directly with model stability or plasticity.

\paragraph{Hyper-parameter search for AdaBOP} AdaBOP requires selecting four hyper-parameters - $svd_{thr} \in [0,1]$, which represents the \% of bottom right singular vectors (after using svd on the task feature space) that are selected as the task null space, $\lambda_s$ and $\lambda_l$, which control the strength of the orthogonality constraint (when projecting new task gradients to previous task null space) (i.e. controls stability) for parameters with high and low correlation to past tasks, respectively, and $\epsilon$, which is the threshold for differentiating high and low correlation. $svd_{thr}$ is set to 0.01, ($\lambda_s$, $\lambda_l$) are selected from a grid search $\in \{(0.1,0.3),(0.03,0.08),(0.005,0.01)\}$, and $\epsilon$ is selected from a grid search $\in \{0.002, 0.004\}$, based on values suggested by the authors, using the current task validation set performance for selection.

\paragraph{DIL specific setups}For DIL, class-weighting is applied to the cross-entropy loss to tackle the heavy class imbalance. Class weights are calculated as the scaled inverse frequency of classes in the full dataset.
\paragraph{CIL specific setups}For CIL with single-head setting, a relaxed-balanced-softmax is used per \cite{JODELET2022RBS} with the suggested default hyper-parameter value.

\begin{table*}[h]
\caption{Task Orders; TG - Toys and Games, MI - Musical Instruments, DM - Digital Music, CDV - CDs and Vinyl, AUTO - Automotive, AIV - Amazon Instant Video; ALC - Alcohol, SM - Smoking, DR - Drug, EX - Exercise, MED - Medicine, ANX - Anxiety; SP0 - [music, quirky, factoid, remove, negate], SP1 - [praise, sendemail, explain, repeat, affirm], SP2 - [radio, confirm, post, definition, dontcare], SP3 - [recipe, podcasts, currency, events, commandstop], SP4 - [createoradd, stock, locations, hue\_lightoff, audiobook], SP5 - [ticket, game, hue\_lightchange, querycontact, likeness], SP6 - [music, sendemail, post, events, audiobook], SP7 - [quirky, explain, definition, commandstop, ticket], SP8 - [factoid, repeat, dontcare, createoradd, game], SP10 - [negate, radio, podcasts, locations, querycontact], SP11 - [praise, confirm, currency, hue\_lightoff, likeness]}
\begin{center}
\begin{tabular}{lllllll}
\toprule
\multicolumn{1}{c}{Dataset}  &\multicolumn{6}{c}{Tasks}
\\
\midrule
\multirow{ 3}{*}{Behaviour} &ALC &SM &DR &EX &MED &ANX \\
&ANX &MED &ALC &EX &SM &DR\\
&DR &EX &MED &SM &ANX &ALC\\
\midrule
\multirow{ 3}{*}{Intent} &SP0 &SP1 &SP2 &SP3 &SP4 &- \\
&SP5 &SP4 &SP2 &SP3 &SP1 &- \\
&SP11 &SP6 &SP10 &SP8 &SP7 &- \\
\bottomrule
\end{tabular}
\end{center}
\label{task-orders}
\end{table*}

\paragraph{Task Ordering} Table \ref{task-orders} shows the sequences of task orders used in experiments for each dataset.

\section{Analysis of Parameter Weighting Strategies with Relative-Importance}\label{section:analysis}

To understand the effect of treating parameters with high and low $\alpha_k^{rel}$ differently, as well as to motivate how $\alpha_k^{rel}$ should be used in the main continual learning phase, we explore three different weighting strategies, as shown in equations \ref{eqn:alpha_kdash_cf} to \ref{eqn:alpha_kdash_ap}. In each of these equations, the new constraint $\alpha_k'$ for each parameter is determined by weighting the original $\alpha_{k-1}$ value using $\alpha_k^{rel}$, where $\tau_{\alpha^{rel}}$ is the cut-off for differentiating high and low relative-importance and  $\lambda_{up}$ and $\lambda_{down}$ are scaling hyper-parameters. Equation \ref{eqn:alpha_kdash_cf} focuses on CF avoidance, where the constraint on parameters with high relative-importance is increased through the weight $\lambda_{up}.\alpha_k^{rel}$, while the constraint on parameters with low relative-importance is unmodified. $\lambda_{up}$ is lower bounded by $1/\tau_{\alpha^{rel}}$ to ensure $\lambda_{up}.\alpha_k^{rel}>1$ (to mitigate CF). Equation \ref{eqn:alpha_kdash_kt} focuses on plasticity and KT, where the constraint on parameters with low relative-importance is decreased through the weight $\lambda_{down}.\alpha_k^{rel}$, while the constraint on parameters with high relative-importance is unmodified. $\lambda_{down}$ is upper bounded by 1 to ensure $\lambda_{down}.\alpha_k^{rel}<1$ (to enable plasticity/KT). Our final proposed approach, Equation \ref{eqn:alpha_kdash_ap}, combines the above two to simultaneously modify the constraint for both sets of parameters.

CF Focus:\\
\begin{equation}
\label{eqn:alpha_kdash_cf}
    \alpha_k' =
    \begin{cases}
        \lambda_{up}.\alpha_k^{rel}.\alpha_{k-1} & \text{if $\alpha_k^{rel}>\tau_{\alpha^{rel}}$} \\
        \alpha_{k-1} & \text{otherwise}
    \end{cases}
\end{equation}

KT Focus:\\
\begin{equation}
\label{eqn:alpha_kdash_kt}
    \alpha_k' =
    \begin{cases}
        \alpha_{k-1} & \text{if $\alpha_k^{rel}>\tau_{\alpha^{rel}}$} \\
        \lambda_{down}.\alpha_k^{rel}.\alpha_{k-1} & \text{otherwise}
    \end{cases}
\end{equation}

CF-KT Balance:\\
\begin{equation}
\label{eqn:alpha_kdash_ap}
    \alpha_k' =
    \begin{cases}
        \lambda_{up}.\alpha_k^{rel}.\alpha_{k-1} & \text{if $\alpha_k^{rel}>\tau_{\alpha^{rel}}$} \\
        \lambda_{down}.\alpha_k^{rel}.\alpha_{k-1} & \text{otherwise}
    \end{cases}
\end{equation}

\begin{figure*}[!h]
    \centering
    \begin{subfigure}[t]{1.0\textwidth}
        \centering
        \includegraphics[height=2.0in]{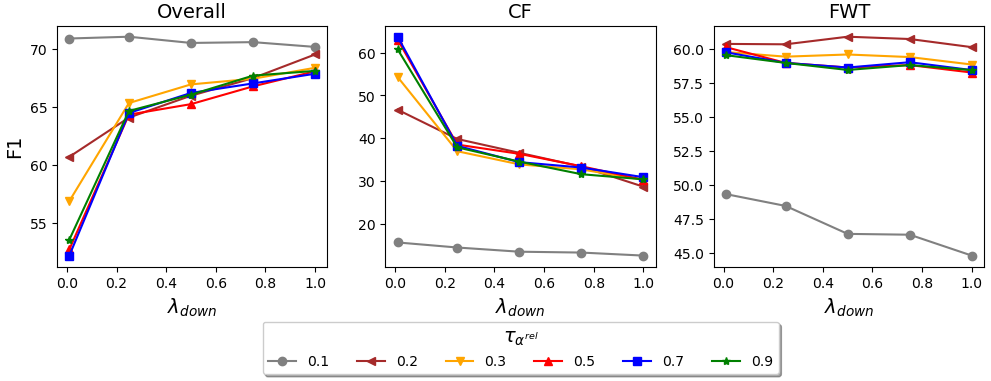}
        \caption{KT-Focus}
        \label{fig:cil_analysis_weightingstrat_and_hyp-a}
    \end{subfigure}
    \begin{subfigure}[t]{1.0\textwidth}
        \centering
        \includegraphics[height=2.0in]{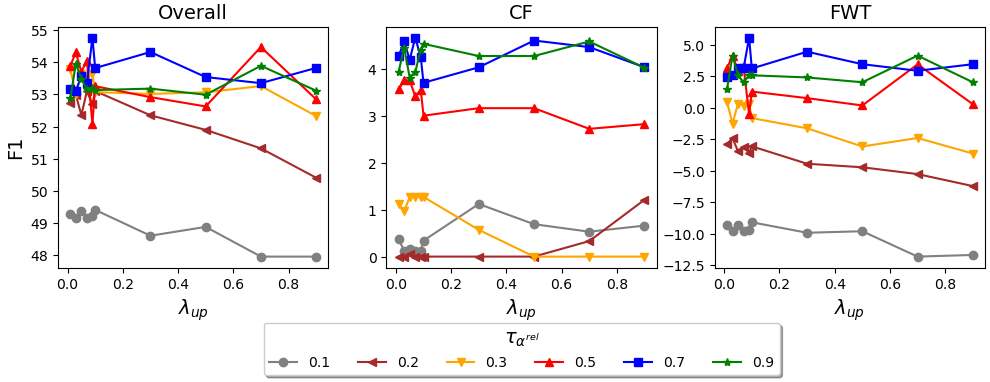}
        \caption{CF-Focus}
        \label{fig:cil_analysis_weightingstrat_and_hyp-b}
    \end{subfigure}
    \caption{Impact of different weighting strategies (CF-Focus, KT-Focus) and look-ahead hyper-parameters on CL performance metrics after learning the second task, under class-incremental learning. 
     $\lambda_{up}$ is scaled by $\lambda_{max}/\lambda$.
    At this step ANCL-MAS results in (CF=4\%, FWT=31\%), and ANCL-LWF in (CF=42\%, FWT=60\%).}
    \label{fig:cil_analysis_weightingstrat_and_hyp_1}
\end{figure*}

\begin{figure*}[!h]
    \centering
    \begin{subfigure}[t]{1.0\textwidth}
        \centering
        \includegraphics[height=2.0in]{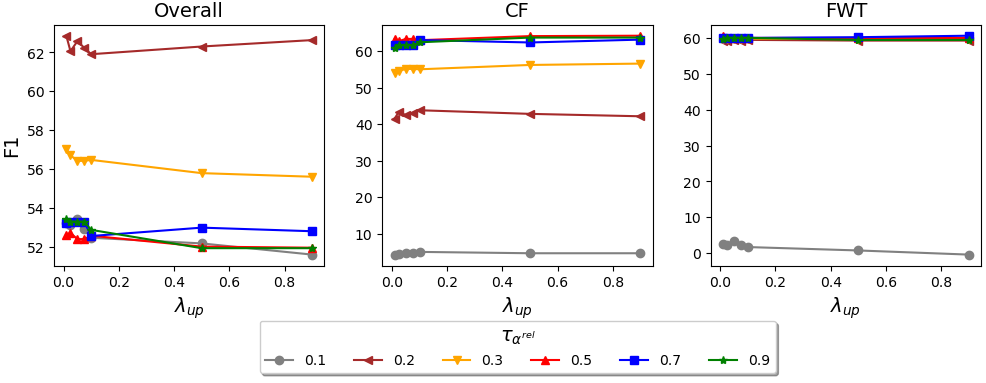}
        \caption{CF-KT Balance: $\lambda_{down}=0.01$}
        \label{fig:cil_analysis_weightingstrat_and_hyp-c}
    \end{subfigure}
    \begin{subfigure}[t]{1.0\textwidth}
        \centering
        \includegraphics[height=2.0in]{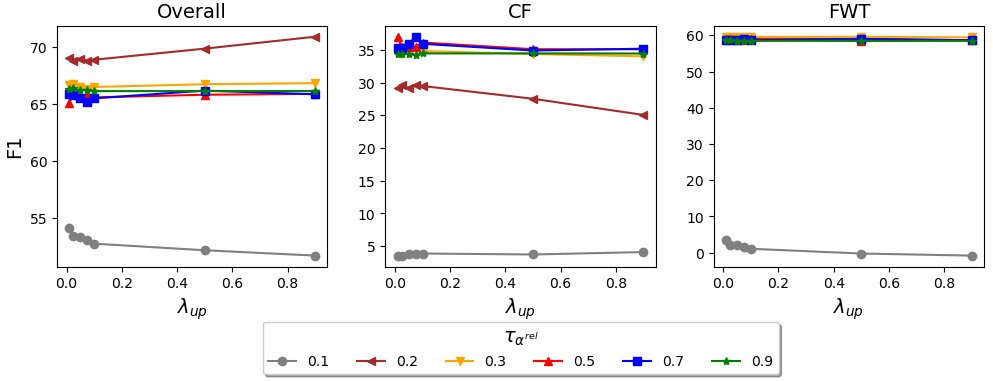}
        \caption{CF-KT Balance: $\lambda_{down}=0.5$}
        \label{fig:cil_analysis_weightingstrat_and_hyp-d}
    \end{subfigure}
    \begin{subfigure}[t]{1.0\textwidth}
        \centering
        \includegraphics[height=2.0in]{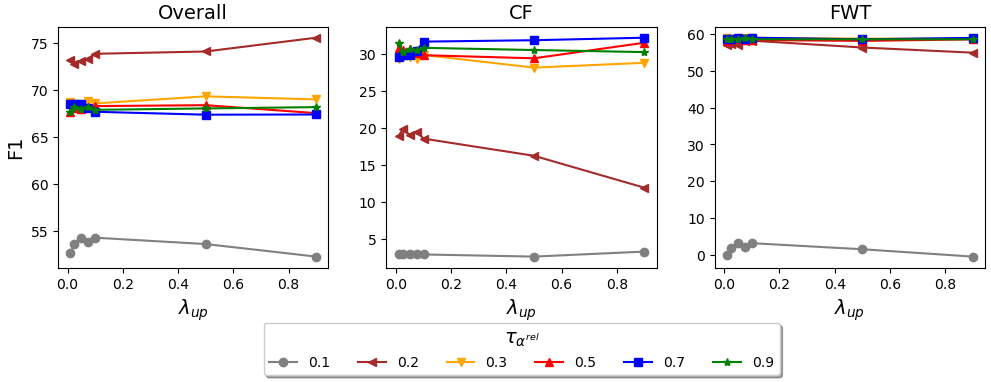}
        \caption{CF-KT Balance: $\lambda_{down}=1$}
        \label{fig:cil_analysis_weightingstrat_and_hyp-e}
    \end{subfigure}
    \caption{Impact of the CF-KT Balance strategy and look-ahead hyper-parameters on CL performance metrics after learning the second task, under class-incremental learning. 
    x-axis is scaled by $\lambda_{max}/\lambda$.
    At this step ANCL-MAS results in (CF=4\%, FWT=31\%), and ANCL-LWF in (CF=42\%, FWT=60\%).}
    \label{fig:cil_analysis_weightingstrat_and_hyp_2}
\end{figure*}

Next, we empirically analyse the three weighting strategies proposed above and various hyper-parameter choices. For this, we consider the second task of the class-incremental sequence. Figures \ref{fig:cil_analysis_weightingstrat_and_hyp_1} and \ref{fig:cil_analysis_weightingstrat_and_hyp_2} show the impact of different weighting strategies and choices of hyper-parameters for $\lambda_{down}$, $\lambda_{up}$ and $\tau_{\alpha^{rel}}$ on the continual learning performance. Crucially, only the LA hyper-parameters are varied, while $\lambda$ remains the same as for the baseline ANCL-MAS.

\paragraph{Stability-plasticity trade-off:} Comparing Figures \ref{fig:cil_analysis_weightingstrat_and_hyp-a} and \ref{fig:cil_analysis_weightingstrat_and_hyp-b}, we see that the KT-Focus formulation results in higher FWT scores indicating better model plasticity, while the CF-Focus formulation results in much lesser CF albeit at the cost of model plasticity. With the CF-KT Balance formulation (Figures \ref{fig:cil_analysis_weightingstrat_and_hyp-c}, \ref{fig:cil_analysis_weightingstrat_and_hyp-d} and \ref{fig:cil_analysis_weightingstrat_and_hyp-e}), we can achieve a good balance between stability and plasticity, under the right hyper-parameter choices. 

\paragraph{Impact of $\tau_{\alpha^{rel}}$:} In general, increasing $\tau_{\alpha^{rel}}$ threshold amounts to relaxing the regularisation constraint on more parameters, for all formulations considered. When setting $\tau_{\alpha^{rel}}=0.1$, effectively fewer parameters can be updated for learning new task data than at higher thresholds and accordingly we see that this setting results in least FWT and CF across all formulations. Increasing $\tau_{\alpha^{rel}}$ beyond $0.3$ often has identical results. To understand this further, we analyse the $\tau_{\alpha^{rel}}$ distribution at various model layers and components.
Interestingly, we find that at most model layers and components, the mean of the distribution falls around $\tau_{\alpha^{rel}}=0.3$. This implies that as we increase the $\tau_{\alpha^{rel}}$ threshold from $0.1$ to $0.3$, the number of parameters with relaxed constraint increases significantly, which explains the significant change in performance. Increasing the threshold further however affects fewer additional parameters, resulting in either similar performance or minor effects. This supports setting $\tau_{\alpha^{rel}}$ to a value slightly lower than the distribution mean to ensure a good stability-plasticity trade-off. An ablation study of different choices is depicted in Table \ref{tab:ablation-tau-alpha} (Appendix \ref{ap:add_results}).

\paragraph{Impact of $\lambda_{down}$:} $\lambda_{down}$ acts only on parameters with low relative-importance (i.e. $\alpha_{k}^{rel}<=\tau_{\alpha^{rel}}$) and decreasing it amounts to relaxing the regularisation constraint on these parameters. 
Figure \ref{fig:cil_analysis_weightingstrat_and_hyp-a} shows that under the KT-Focus formulation, when $\tau_{\alpha^{rel}}>=0.3$, decreasing $\lambda_{down}$ at first steadily increases CF (up to 40\%) until at very low $\lambda_{down}$ (=0.01), there is a drastic increase in CF (up to 60\%), while throughout FWT remains the same ($\sim$60\%). 
For $\tau_{\alpha^{rel}}=0.2$, the increase in CF is less drastic. Then, for $\tau_{\alpha^{rel}}=0.1$, decreasing $\lambda_{down}$ only minimally impacts CF (which remains at $<$30\%), while FWT steadily improves (upto 50\%).
This validates our hypothesis that parameters with low relative-importance ($\tau_{\alpha^{rel}}<0.3$ in this case) contribute less to forgetting of past-task knowledge and relaxing the regularisation constraint on these can help with model plasticity. 
In the CF-KT Balance formulation (figure \ref{fig:cil_analysis_weightingstrat_and_hyp_2}), parameters with high relative-importance are additionally given increased regularisation. For $\tau_{\alpha^{rel}}>=0.3$, decreasing $\lambda_{down}$ has similar trends as the KT-Focus formulation. 
For $\tau_{\alpha^{rel}}<0.3$, at all $\lambda_{down}$, CF is reduced further compared to the KT-Focus formulation, while still maintaining high FWT  ($\sim$60\%) when $\tau_{\alpha^{rel}}$ is not too low (=0.2). This further validates our relative-importance hypothesis and indicates that the CF-KT Balance formulation achieves superior stability-plasticity trade-off. 

\paragraph{Impact of $\lambda_{up}$:} $\lambda_{up}$ acts only on parameters with high relative-importance (i.e. $\alpha_{k}^{rel}>\tau_{\alpha^{rel}}$) and increasing it amounts to stricter constraints on these parameters. 
We see that in the CF-KT Balance formulation (figures \ref{fig:cil_analysis_weightingstrat_and_hyp_2}), increasing $\lambda_{up}$ can help reduce CF with minimal impact to FWT.

\paragraph{Conclusion:} The above analysis provides an indication of the impact of different hyper-parameters on the continual learning performance, while also validating the assumptions made with respect to high and low relative-importance. Moreover, the CF-KT Balance formulation is found to be the most promising approach.
Based on this analysis, the following approach to setting the three LA hyper-parameters is followed, as described in Appendix \ref{ap:alg} - (1) $\tau_{\alpha^{rel}}$ is set slightly lower than the distribution mean at a given model component and layer, ensuring a good balance between stability and plasticity. An ablation study of different choices is conducted in Appendix \ref{ap:ablation-tau-alpha-rel}. (2) $\lambda_{down}$ and $\lambda_{up}$ are set to their maximum values of 1 and $\lambda_{max}/\lambda$, respectively. $\lambda_{down}$ is decreased first, ensuring minimum forgetting on previous tasks. (3) When further decrease in $\lambda_{down}$ results in a performance plateau, $\lambda_{up}$ is decreased next.

\section{Additional Results}\label{ap:add_results}

\subsection{Comparison of Stability-Plasticity Trade-Off and Backward Transfer Across Thresholds}\label{ap:thresholds-analysis}

\begin{table}[t]
\centering
\caption{Comparison of Stability-Plasticity Trade-Off and Backward Transfer: CL performance at the end of all tasks in the sequence, at different thresholds, measured in F1 scores. Best is indicated in bold. $\uparrow$ indicates higher is better, $\downarrow$ indicates lower is better. Results on the second class-incremental sequence.}\label{tab:analysis-across-thresholds}
\begin{tabular}{cclllll}
\toprule
$thr$ &$\lambda$ & Method &{Ov$\uparrow$} &{CF$\downarrow$} &{BWT+$\uparrow$} &{FWT$\uparrow$}\\
\midrule
90 &0.21 & ANCL-LWF & 25.09 & 83.46 & 0.00 &\textbf{22.89}\\
90 &4.5 & ANCL-MAS & \textbf{36.98} & \textbf{67.95} & \textbf{0.33} &22.23\\
90 &4.5 & LA-MAS & 32.91 & 71.37 & 0.00 &20.57\\
\midrule
80 &0.22 & ANCL-LWF & 25.09 & 83.46 & 0.00 &\textbf{22.89}\\
80 &6.17 & ANCL-MAS & \textbf{52.62} & \textbf{45.34} & 0.00 &19.17\\
80 &6.17 & LA-MAS & 48.92 & 52.08 & \textbf{11.36} &21.30\\
\midrule
70 &0.46 & ANCL-LWF & 24.75 & 69.55 & 3.64 &8.54\\
70 &6.85 & ANCL-MAS & 45.50 & \textbf{36.58} & 4.31 &1.52\\
70 &6.85 & LA-MAS & \textbf{50.03} & 46.94 & \textbf{7.80} &\textbf{17.54}\\
\midrule
60 &0.51 & ANCL-LWF & 29.95 & 73.14 & 0.00 &18.64\\
60 &21.83 & ANCL-MAS & 45.45 & \textbf{44.39} & 0.00 &9.26\\
60 &21.83 & LA-MAS & \textbf{45.84} & 54.58 & \textbf{3.05} &\textbf{19.94}\\
\bottomrule
\end{tabular}
\end{table}

Table \ref{tab:analysis-across-thresholds} depicts the continual learning performance of the LA and ANCL methods as the threshold is varied. The results indicate that ANCL-LWF in general struggles to balance stability and plasticity, with comparatively high forgetting at all considered thresholds, resulting in consistently poorer overall performance than ANCL-MAS and LA-MAS. Interestingly, at very high threshold (=90), forgetting is significantly high with negligible to no backward knowledge transfer, while lowering the threshold (=70) increases \textit{both} stability and backward knowledge transfer for all methods. This may indicate that to achieve backward knowledge transfer, some model stability is also desirable. In other words, at high threshold, when regularisation ($\lambda$) is very low, model parameters can undergo significant changes to accommodate new task knowledge, causing forgetting rather than backward transfer. When regularisation is moderate and model parameters are not significantly updated, backward transfer becomes feasible. As the threshold is reduced further (=60), however, high regularisation affects backward transfer. LA-MAS, by allowing parameters with low relative importance to be less regularised even when they have high past-task importance, achieves better backward transfer and maintains high plasticity even at lower thresholds, at the cost of some model stability.

\subsection{Ablation of $\tau_{\alpha^{rel}}$ choice}\label{ap:ablation-tau-alpha-rel}
Table \ref{tab:ablation-tau-alpha} shows the stability-plasticity trade-off as $\tau_{\alpha^{rel}}$ is reduced, as new tasks are introduced. 
Reducing $\tau_{\alpha^{rel}}$ effectively reduces the number of parameters that are allowed to be easily updated. When learning the second task, reducing $\tau_{\alpha^{rel}}$ steadily helps reduce forgetting, resulting in better stability-plasticity trade-off. When learning subsequent tasks, network capacity begins to saturate. Therefore, while reducing $\tau_{\alpha^{rel}}$ helps reduce forgetting initially, at lower $\tau_{\alpha^{rel}}$ values, regularisation also needs to be lowered (i.e. lower $\lambda_{down}$, $\lambda_{up}$) to achieve comparable plasticity. This in turn results in increased forgetting at lower $\tau_{\alpha^{rel}}$ values. A good trade-off is achieved at intermediate $\tau_{\alpha^{rel}}$ values of 0.8 to 0.85.

\begin{table}[t!]
\centering
\caption{Analysing the effect of $\tau_{\alpha^{rel}}$ choice on the continual learning performance. Best is indicated in bold. Results on the second class-incremental learning sequence.}\label{tab:ablation-tau-alpha}
\begin{tabular}{ccccclll}
\toprule
Tasks & $\tau_{\alpha^{rel}}$ & $\lambda$ & $\lambda_{down}$ & $\lambda_{up}$ & Ov & CF & FWT \\
&($\times$ distribution mean) & & &($\times$ $\lambda_{max}/\lambda$) \\
\midrule
2 & 0.90 & 4.50 & 1.00 & 1.00 & 70.56 & 25.40 & \textbf{58.29} \\
2 & 0.85 & 4.50 & 1.00 & 1.00 & 71.34 & 23.84 & \textbf{58.29} \\
2 & 0.80 & 4.50 & 1.00 & 1.00 & 71.8 & 22.54 & 57.91 \\
2 & 0.75 & 4.50 & 1.00 & 1.00 & 71.56 & 22.63 & 57.51 \\
2 & 0.70 & 4.50 & 1.00 & 1.00 & \textbf{72.48} & \textbf{20.19} & 56.93 \\
\midrule
3 & 0.90 & 4.50 & 0.80 & 1.00 & 59.42 & 42.84 & 31.28 \\
3 & 0.85 & 4.50 & 0.73 & 1.00 & \textbf{61.57} & \textbf{39.41} & 31.07 \\
3 & 0.80 & 4.50 & 0.59 & 1.00 & 59.96 & 42.18 & 31.43 \\
3 & 0.75 & 4.50 & 0.20 & 0.39 & 58.7 & 44.11 & \textbf{31.48} \\
3 & 0.70 & 4.50 & 0.20 & 1.00 & 59.17 & 42.68 & 30.74 \\
\midrule
4 & 0.90 & 4.50 & 0.33 & 1.00 & 50.35 & 53.22 & 20.73 \\
4 & 0.85 & 4.50 & 0.32 & 1.00 & 50.84 & 52.30 & 20.46 \\
4 & 0.80 & 4.50 & 0.31 & 1.00 & \textbf{53.16} & \textbf{49.03} & 20.27 \\
4 & 0.75 & 4.50 & 0.06 & 1.00 & 50.45 & 53.16 & \textbf{20.79} \\
4 & 0.70 & 4.50 & 0.01 & 0.58 & 47.86 & 55.40 & 19.59 \\
\bottomrule
\end{tabular}
\end{table}

\section{Computation Cost}\label{ap:comp-cost}
Table \ref{tab:comp-cost} provides the computation cost estimate for each of the compared continual learning methods in terms of the number of forward and backward passes through the data, ignoring the cost of hyper-parameter search. UPGD, designed for online learning, computes the parameter-utility at each update step using the gradient at that step and thus requires the same number of passes through the data as SEQ. AdaBOP, which uses the gradient space to estimate task correlations, requires an additional pass through the data at the start of training each task, to collect the gradients for correlation estimation. RP2F estimates parameter importance at each epoch. Notably, RP2F with fisher parameter importance is more efficient (comparable to ANCL-MAS, LA-MAS) than perturbation based importance, which requires separate passes through the data for each parameter being perturbed at each epoch. LA-MAS and ANCL-MAS have similar computation cost, where the overhead over SEQ consists of parameter importance estimation at the end of each task and auxiliary/LA network training. ANCL-LWF can be more costly as the data is passed through both frozen and current network during MCL.

\begin{table}[t!]
\caption[Computation Cost]{Computation Cost: Number of forward and backward passes through the data when training on a sequence of two tasks, $b_1$ and $b_2$ represent the number of batches in the first and second task, respectively, $E$ denotes the number of epochs taken to converge in each training phase and $N_p$ denotes the number of model parameters perturbed.} 
\label{tab:comp-cost}
\begin{center}
\begin{tabular}{cl} 
\toprule
{Method} &{\# Passes}\\ 
\midrule
MTL &$E.(b_1 + b_2)$
\\ SEQ &$E_1.b_1 + E_2.b_2$
\\ UPGD &$E_1.b_1 + E_2.b_2$ 
\\ AdaBOP &$b_1 + E_1.b_1 + b_2 + E_2^{mcl}.b_2$
\\ RP2F-fisher &$E_1.b_1 + b_1 + E_2^{mcl}.(b_2 + b_2)$
\\ LA-MAS &$E_1.b_1 + b_1 + E_2^{aux}.b_2 + b_2 + E_2^{mcl}.b_2$
\\ ANCL-MAS &$E_1.b_1 + b_1 + E_2^{aux}.b_2 + b_2 + E_2^{mcl}.b_2$
\\ ANCL-LWF &$E_1.b_1 + E_2^{aux}.b_2 + E_2^{mcl}.(b_2 + b_2)$
\\ RP2F-perturb &$E_1.b_1 + b_1 + E_2^{mcl}.(b_2 + N_p.b_2)$
\\ \bottomrule
\end{tabular}
\end{center}
\end{table}

\end{document}